\documentclass[a4paper,fleqn]{cas-sc}

\usepackage[numbers,square,sort&compress]{natbib}
\usepackage{cite}
\usepackage{amsmath,amssymb,amsfonts}
\usepackage{algorithmic}
\usepackage{graphicx}
\usepackage{textcomp}
\usepackage{xcolor}
\def\BibTeX{{\rm B\kern-.05em{\sc i\kern-.025em b}\kern-.08em
    T\kern-.1667em\lower.7ex\hbox{E}\kern-.125emX}}
\usepackage{hyperref}
\usepackage{tabularx}
\usepackage{comment}
\usepackage{multicol}
\usepackage{multirow}
\usepackage{placeins}
\usepackage{graphicx}
\usepackage{booktabs}
\usepackage{subcaption}
\usepackage{caption}

\usepackage{url}
\usepackage{cleveref}
\crefname{appendix}{Appendix}{Appendices}
\crefname{table}{Table}{Tables}
\crefname{figure}{Figure}{Figures}

\newcolumntype{Y}{>{\centering\arraybackslash}X}
\newcolumntype{Z}{>{\centering\arraybackslash}p{2.2cm}}
\newcolumntype{W}{>{\centering\arraybackslash}p{3cm}}
\newcolumntype{V}{>{\centering\arraybackslash}p{1.0cm}}
\newcolumntype{U}{>{\centering\arraybackslash}p{5cm}}

\def\tsc#1{\csdef{#1}{\textsc{\lowercase{#1}}\xspace}}
\tsc{WGM}
\tsc{QE}
\tsc{EP}
\tsc{PMS}
\tsc{BEC}
\tsc{DE}

\begin{document}
\let\WriteBookmarks\relax
\def\floatpagepagefraction{1}
\def\textpagefraction{.001}
\shorttitle{Masked Conditioning for Deep Generative Models}
\shortauthors{P. Mueller et~al.}

\title [mode = title]{Masked Conditioning for Deep Generative Models}

\author[1,2]{Phillip Mueller}[orcid=0009-0007-4612-3224]
\cormark[1]
\fnmark[1]
\ead{phillip.mueller@bmw.de}

\credit{Conceptualization, Methodology, Formal analysis, Writing - original draft, Software, Visualization, Project administration}

\author[1,3]{Jannik Wiese}
\credit{Methodology, Software, Writing - review \& editing}

\author[1]{Sebastian Mueller}
\credit{Conceptualization, Writing - review \& editing}

\author[2]{Lars Mikelsons}
\credit{Writing - review \& editing, Supervision}

\affiliation[1]{organization={BMW Group},
            addressline={Knorrstrasse 147}, 
            city={Munich},
            postcode={80788}, 
            state={Bavaria},
            country={Germany}}
            
\affiliation[2]{organization={University of Augsburg},
            addressline={Am Technologiezentrum 8}, 
            city={Augsburg},
            postcode={86159}, 
            state={Bavaria},
            country={Germany}}

\affiliation[3]{organization={Ludwig-Maximilians-University Munich},
            addressline={Geschwister-Scholl-Platz 1}, 
            city={Munich},
            postcode={80539}, 
            state={Bavaria},
            country={Germany}}
            
\cortext[cor1]{Corresponding author}

\begin{abstract}
Datasets in engineering domains are often small, sparsely labeled, and contain numerical as well as categorical conditions. Additionally. computational resources are typically limited in practical applications which hinders the adoption of generative models for engineering tasks.
We introduce a novel masked-conditioning approach, that enables generative models to work with sparse, mixed-type data. We mask conditions during training to simulate sparse conditions at inference time. For this purpose, we explore the use of various sparsity schedules that show different strengths and weaknesses. In addition, we introduce a flexible embedding that deals with categorical as well as numerical conditions. We integrate our method into an efficient variational autoencoder as well as a latent diffusion model and demonstrate the applicability of our approach on two engineering-related datasets of 2D point clouds and images.
Finally, we show that small models trained on limited data can be coupled with large pretrained foundation models to improve generation quality while retaining the controllability induced by our conditioning scheme.
\end{abstract}

\begin{keywords}
Deep Generative Models \sep Latent Diffusion Models \sep Conditioning \sep Variational Autoencoders
\end{keywords}

\maketitle

\section{Introduction} Deep Generative Models (DGMs) have demonstrated great success across a wide range of tasks, including image generation and natural language processing. However, the application of DGMs in engineering design remains limited. As of now, Generative Adversarial Networks (GANs) are mainly used with low-fidelity design data like airfoil profiles~\citep{nobariPcDGANContinuousConditional2021,chenPaDGANLearningGenerate2021,chenMOPaDGANReparameterizingEngineering2021}. But are also applied to design-related image data \citep{nobariCreativeGANEditingGenerative2021, karras2020analyzingimprovingimagequality, karras2021aliasfreegenerativeadversarialnetworks}. Variational Autoencoders (VAEs) are proposed for 3D-shapes~\citep{zhang3DShapeSynthesis2019} with limited complexity.
However, state-of-the-art visual DGMs, namely Diffusion- and Transformer-based approaches, have not found widespread application in engineering design, although their potential is known \citep{alamAutomationAugmentationRedefining2024,MuellerExploringthePotentials2024, poStateArtDiffusion2023}. Several critical challenges inhibit their broader adoption in this domain. A primary obstacle is the requirement of large-scale datasets for training~\citep{alamAutomationAugmentationRedefining2024,picardConceptManufacturingEvaluating2023,berthelotUnderstandingImprovingInterpolation2018}. General purpose DGMs are typically trained and conditioned on vast, internet-scale datasets with pairs of images and text-descriptions~\citep{poStateArtDiffusion2023, rombachHighResolutionImageSynthesis2022}. Such quantities are typically not available in engineering design applications. In reality, the case-specific datasets are often small, sparse, and contain a mixture of categorical and numerical data. An example of a realistic dataset with these properties is the DVM-Car dataset \citep{Huangdvmcar2022}.  This difference in data scale and quality necessitates adaptation of standard generative modeling techniques for increased efficiency.
In addition to data limitations, the computational cost associated with training DGMs presents another challenge. Large models demand substantial computing resources. For example, training Stable Diffusion from scratch initially took 150k A100 hours \citep{rombachHighResolutionImageSynthesis2022}. The utilized dataset (LAION-5B) contains 5 billion pairs of images and captions \citep{schuhmann2022laion5bopenlargescaledataset}. Efforts of this scale are impractical for engineering applications where the datasets are specialized and significantly smaller, and computational ressources are limited.


To address these challenges and make DGMs more applicable for engineering design, we propose a flexible conditioning architecture that can handle real-world design data. Specifically, we target the problem of incomplete conditioning datasets, sparsely annotated data, and mixed data types. We investigate methods to achieve robust model generalization with minimal data through sparsity scheduling when training models on incomplete annotations. 
We demonstrate the applicability of our approach across different types of engineering design data and apply it to conditional VAEs trained on point-cloud data, Latent Diffusion Models (LDMs) for CAD-like and sketch images, as well as LDMs for natural image data. We put specific emphasis on training efficiency. This is based on our results, revealing that we can combine small-scale DGMs, trained on engineering design data, with large-scale pretrained models like Stable Diffusion \citep{rombachHighResolutionImageSynthesis2022}, bootstrapping their powerful image priors to increase image quality and realism.
Such a two-stage approach could leverage recent works such as InsertDiffusion \citep{MuellerInsertDiffusion2024} or refinement in Stable Diffusion XL \citep{podellSDXLImprovingLatent2023}.

\section{Background} Our work introduces a novel architecture to condition DGMs with sparse and mixed conditioning data. In this section, we review previous works addressing the problem of handling incomplete and sparse data in DGMs and give a brief introduction into conditional VAEs and LDMs, as we later augment these architectures with our conditioning approach.

\subsection{DGMs with Sparse Data}
To deal with sparse inputs in the context of machine learning, several imputation approaches to fill in missing samples exist. 
Training DGMs on incomplete or sparse input (not conditioning) data has previously been addressed in a number of works. \citep{NAZABALhandlingmissingdata2020} propose a framework that allows VAEs to handle datasets with missing values by incorporating a probabilistic treatment of incomplete data during the learning process. Similarly, \citep{collier2021vaespresencemissingdata} develop methods for VAEs to operate in the presence of missing data, enabling the model to predict the missing parts based on the observed data. Aiming to increase applicability on real-world data, \citep{ma2020vaemdeepgenerativemodel} propose a VAE architecture trained in two stages to handle heterogeneous data types.
Working on VAEs applied in the image space, \citep{ivanov2018variational} train the model on sparse image data with masked regions. The VAE is tasked with filling in these regions, effectively learning to generate plausible content for missing parts of the input. The approach demonstrates the potential of VAEs to handle sparse input data, although in a context slightly different from engineering design.
While interesting, these methods do not focus on sparse conditioning data.
Some GAN-based approaches have been proposed that utilize adversarial training to fill in gaps in the conditioning data \citep{odena2016semisupervisedlearninggenerativeadversarial, springenberg2016unsupervisedsemisupervisedlearningcategorical}. However, adversarial training is known to be instable and training an additional model to impute conditions is expensive if a different generative model is supposed to be used downstream.
\subsection{Conditional Deep Generative Models}
\noindent\textbf{Conditional Variational Autoencoders.}
In a classic VAE, the encoder learns to map the input data $x$ to a latent representation $z$. The latent is usually modeled as a Gaussian distribution $q_{\phi}(z|x)=\mathcal{N}{(z;\mu_{\phi}(x),\sigma^2_{\phi}(x))}$ \citep{kingmaAutoEncodingVariationalBayes2013,Kingma_2019_IntroVAE}. The decoder is trained to reconstruct the data from the latent distribution i.e. modelling $p_{\theta}(x|z)$. The training objective is to maximize the Evidence Lower Bound (ELBO) by balancing the reconstruction loss ${\mathbb{E}_{q_{\phi}(z|x)}[\log{p_{\theta}(x|z)}]}$ with the regularization term $KL(q_{\phi}(z|x){\parallel}p(z))$ the latter of which aligns the latent distribution with a known, simple Gaussian prior distribution \citep{kingmaAutoEncodingVariationalBayes2013,Kingma_2019_IntroVAE}.
Conditional VAEs (cVAE) are designed to process additional conditioning information within the generative process. A conditioning variable $c$ is included in the encoder $q_{\theta}(z|x,c)$ and decoder $p_{\theta}(x|z,c)$ \citep{sohnlearningstructuredoutputrepresentations2015, burnapEstimatingExploringProduct2016, yonekuraGeneratingVariousAirfoil2021}. The training objective for cVAE is to maximize the ELBO on the marginal likelihood of the data, conditioned on $c$:
\begin{equation}
\label{eq:cvae}
\footnotesize
  \mathcal{L}(\phi,\theta;x,c)=\mathbb{E}_{q_{\phi}(z|x,c)}[\log p_{\theta}(x|z,c)]-KL(q_{\theta}(z|x,c){\parallel}p(z|c)).  
\end{equation}
 \noindent\textbf{Conditional Latent Diffusion Models.}
 Diffusion models are based on the idea of iteratively destroying the information in the data by adding Gaussian noise and then learning a neural network to reverse this process (denoising) \citep{sohnlearningstructuredoutputrepresentations2015,ho2020denoisingdiffusionprobabilisticmodels,song2022denoisingdiffusionimplicitmodels}. A U-Net-style model \citep{Ronneberger2015UNetCN} is trained to predict the added noise in each timestep, which allows the model to iteratively subtract noise in inference, slowly turning a noise sample into new data. The simplified loss function as proposed by \citep{ho2020denoisingdiffusionprobabilisticmodels} is $\mathcal{L}(\theta)=\mathbb{E}_{t,x_0,\epsilon}[{\parallel}\epsilon-\epsilon_{\theta}(x_t,t){\parallel}^2_2]$, where $\epsilon_{\theta}$ is the model that predicts the noise $\epsilon$ added in each timestep $t$ \citep{princeUnderstandingDeepLearning2023}.
Generating high-resolution images with diffusion models is impractical due to excessive memory requirements \citep{poStateArtDiffusion2023}. Latent diffusion models (LDMs) like Stable Diffusion have been proposed to address this challenge \citep{rombachHighResolutionImageSynthesis2022}. They operate in a latent space instead of the pixel space and therefore consist of a two-stage architecture. A first-stage VAE is trained to compresses and subsequently reconstructs the image. The denoising model sits in between encoder and decoder and operates on only on the compressed latent representation of the data. The Stable Diffusion architecture builds upon the U-Net model and incorporates self-attention and cross-attention blocks \citep{rombachHighResolutionImageSynthesis2022,poStateArtDiffusion2023,vaswaniattention2017}.
The simplest method for injecting additional conditioning information is concatenation of the condition with the intermediate denoising targets \citep{dhariwaldiffusion2021,sahariapalette2022}. It is suitable for a variety of conditioning modalities like reference images and inpainting masks \citep{sahariapalette2022}. A more flexible method is to inject the conditioning signal via cross-attention. In Stable Diffusion, the conditioning information $c$ is processed by a domain-specific encoder $\tau$. This projection $\tau(c)$ is then injected into the intermediate layers of the U-Net via cross-attention \citep{rombachHighResolutionImageSynthesis2022,poStateArtDiffusion2023}. 

\section{Methodology}
\subsection{Embedding of Conditions}
Before the data is introduced into the generative process, we perform the same preprocessing of the data regardless of the model. We differentiate between categorical conditions $y_{\text{cat}} \in \mathbb{N}^{k_{\text{cat}}}$ and numerical conditions $y_{\text{num}} \in \mathbb{R}^{k_{\text{num}}}$. 
All conditions are embedded before being injected into the generative module. Categorical conditions $y_{\text{cat},i}$ are processed through learnable embedding matrix $E_{\text{cat},i}$ which maps each condition to a fixed-length vector $e_{\text{cat},i}$, therefore:
\begin{equation}
\footnotesize
\label{cat_conditions}
    e_{\text{cat},i} = E_{\text{cat},i}\left( y_{\text{cat},i} \right), \quad e_{\text{cat},i} \in \mathbb{R}^{d_{\text{cat}}}.
\end{equation}
For each categorical condition, the embedding matrix is $E_{\text{cat},i} \in \mathbb{R}^{{n_i+1}\times d_{\text{cat}}}$ and maps the condition to an embedding space, where $n_i$ is the number of unique categories plus one to account for the case where the condition is masked.
Numerical conditions $y_{\text{num},j}$ are projected into fixed-length vectors using a learned positional encoding, implemented as a single linear layer $E_{\text{num},j}$. The projected numerical conditions are given by:
\begin{equation}
\footnotesize
\label{num_conditions}
e_{\text{num},j} = E_{\text{num},j}\left( y_{\text{num},j} \right), \quad e_{\text{num},j} \in \mathbb{R}^{d_{\text{num}, i}}.
\end{equation}
Finally, the embedded categorical and numerical conditions are concatenated to form the complete conditioning vector $e_y=[e_{\text{cat}},e_{\text{num}}]$, which is then passed to the generative model.

\subsection{Masking}
Prior to embedding the conditions, we apply masking to the conditional information to simulate missing values, allowing the generation process to condition on an arbitrary subset of available inputs at inference time. We choose two different approaches to masking based on the type of condition. For categorical conditions we reserve a token for missing conditions. For numerical conditions we set the value to -1, since all numerical variables are strictly positive in our dataset.

To enable the model to deal with different levels of sparsity, i.e. produce reasonable results with an arbitrary amount of given conditions, we design a novel training procedure inspired by curriculum learning \citep{bengio_curriculum_2009}. For this purpose, we implement a sparsity scheduler that gives a sparsity level $p_t \in [0,1]$ for each gradient update. Therefore, our sparsity schedule defines a function $f: [0, T] \rightarrow [0,1]$, where $T$ is the number of total training steps we make. We mask each condition with probability $p_t$, i.e. for each condition $c$ the model receives $c$ with probability $1-p_t$ and receives a placeholder value (the reserved token or -1) with probability $p_t$.
We implement a range of sparsity schedules with different behaviors. The \textbf{constant sparsity} schedule returns the same level of sparsity for each gradient update. The \textbf{step sparsity} schedule divides the training run into $N$ segments. For each segment $i$ the sparsity is set to:
\begin{equation}
f_{step}(i) = p_{start} + (i-1) \frac{p_{end} - p_{start}}{N}.
\end{equation}
The \textbf{linear sparsity} schedule performs linear interpolation between start and end sparsity.
The \textbf{exponential sparsity} schedule is implemented as: 
\begin{equation}
f_{exp}(t) = p_{start} + (p_{end} - p_{start}) (1-e^{-t/T}).
\end{equation}
Each sparsity schedule can be run with increasing or decreasing sparsity and can have an arbitrary start and final sparsity level as long as both values are within the $[0,1]$.

\subsection{Masked Conditioning for VAE}
\label{subsec:mcVAE}

For VAE architectures, we propose to use a classical unconditioned encoder $q_{\phi}(z|x)$ to map the data $x$ to a latent representation $z \in \mathbb{R}^{d_z}$ modeled as a Gaussian distribution. This design diverges from conventional conditional VAE (cVAE) architectures, which typically incorporate conditioning data directly into the encoder. 
The rationale behind this design is to allow the decoder to be flexibly guided by the available conditioning information. Incorporating the sparse conditions into the encoder would introduce ambiguity because the latent representation $z$ would already reflect some degree of conditioning. Adding further conditioning information at the decoding stage could then result in inconsistencies, as the sampled latent variable might not fully align with the additional conditioning.
The decision is also inspired by \citep{chira2022imagesuperresolutiondeepvariational,oord2016conditionalimagegenerationpixelcnn}.

In our model, the latent variable $z$ is concatenated with the embedded conditioning vector $e_y$ forming $z_{c}=[z;e_y]$ where $z_{c} \in \mathbb{R}^{d_z+d_y}$. The decoder $p_{\theta}(x|z_c)$ reconstructs the  data, yielding $\hat{x} \sim p_{\theta}(x|z_c)$. The conditional VAE is trained by maximizing the ELBO through:
\begin{equation}
\footnotesize
   \mathcal{L}(\phi,\theta,E)= \mathbb{E}_{q_{\phi(z|x)}}[\log p_{\theta}(x|z_c)] - KL(q_{\phi}(z|x)\parallel p(z)).
\end{equation}
Notably, only the reconstruction term incorporates the embedded and concatenated conditioning information. The Kullback-Leibler Divergence (KL) is calculated solely on the unconditioned latent variable $z$, ensuring that the encoder learns a smooth and regularized projection of the data. The conditioning information is known, and thus does not require regularization.

\begin{figure*}
    \begin{subfigure}{0.48\textwidth}
        \centering
        \includegraphics[width=0.97\linewidth]{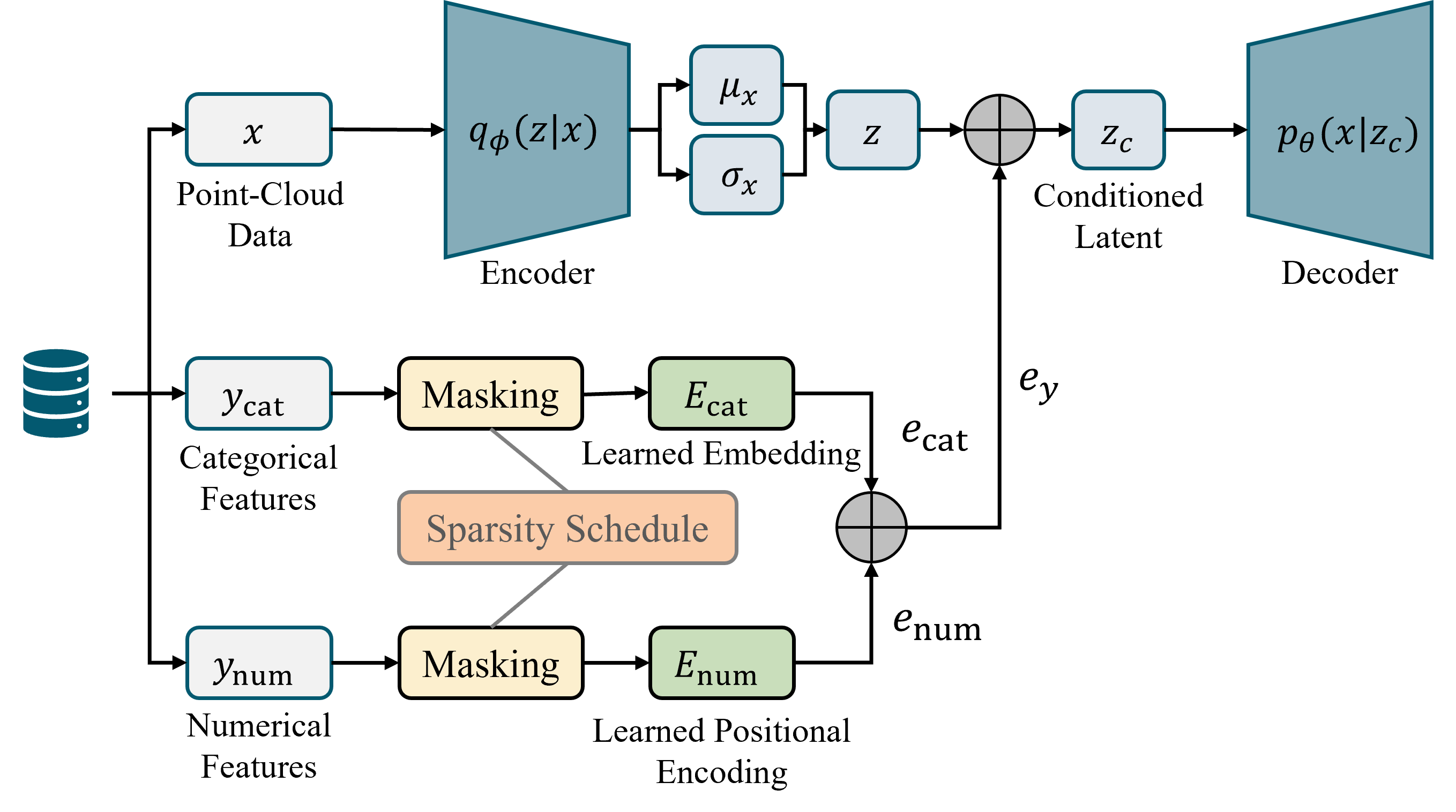}
    \end{subfigure}
    \begin{subfigure}{0.48\textwidth}
        \centering
        \includegraphics[width=0.97\linewidth]{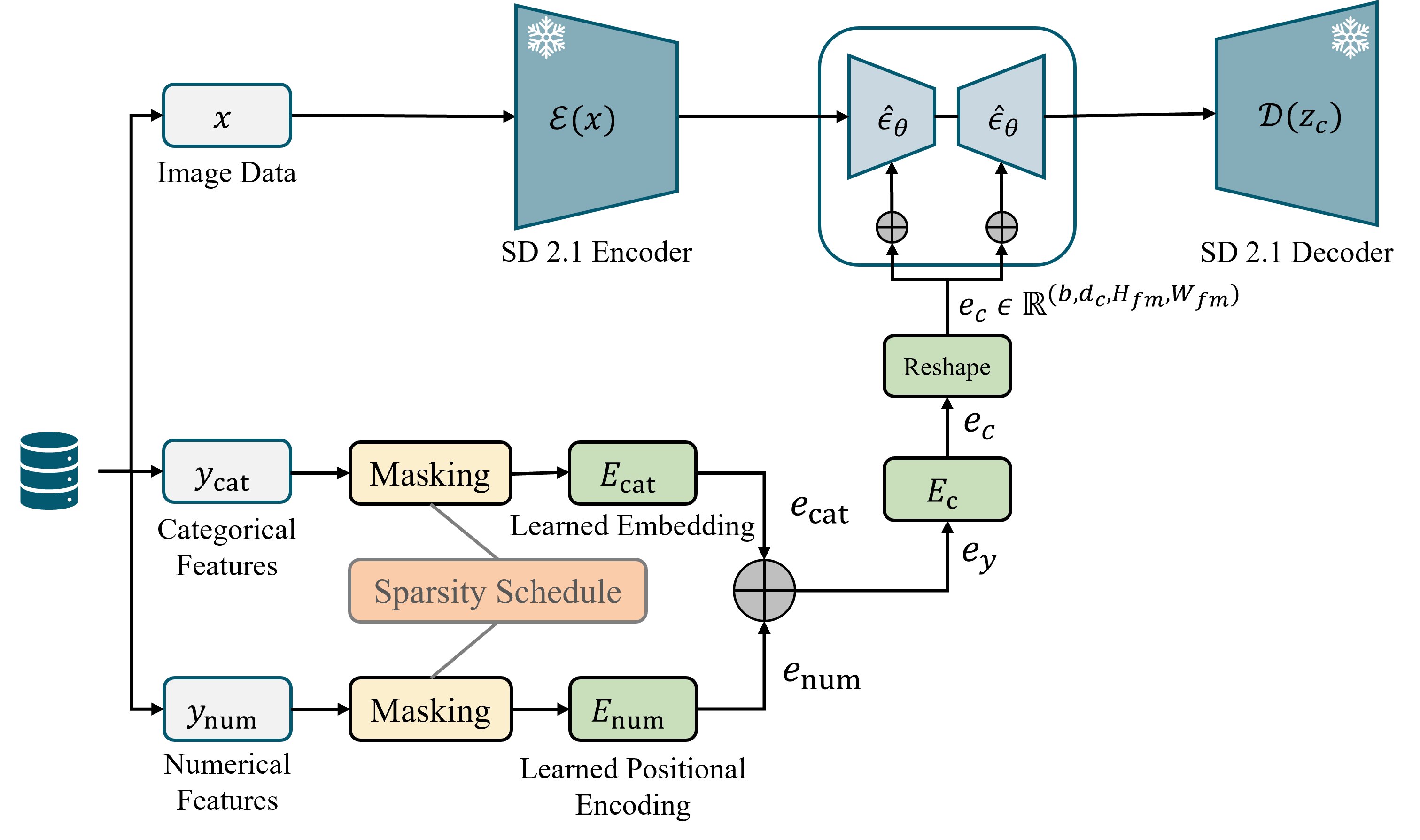}
    \end{subfigure}
    \caption{\textbf{Left:} Architecture of our masked conditioning approach, applied to a VAE. \textbf{Right:} Architecture of our masked conditioning approach applied to a diffusion model.}
    \label{fig:Architecture}
\end{figure*}

\subsection{Masked Conditioning for LDM}
\label{subsec:mcLDM}
The conditioning mechanism for LDMs closely mirrors the approach used for the cVAE. The numerical and categorical conditions are embedded and concatenated as described in the section above. The key difference for training LDMs with our conditioning is the injection of the embedded conditions into the generative process. They are concatenated with the feature maps that are passed to the first ResNet block at each resolution level in the U-Net backbone of the LDM (see \Cref{fig:Architecture}).

At each Unet resolution, the embedded and concatenated conditioning vector $e_y=[e_{\text{cat}},e_{\text{num}}]\in\mathbb{R}^{d_y}$ is passed through a learnable linear layer to reduce its dimensionality to produce the smaller combined embedding $e_c = E_c(e_y)$ with $e_c\in\mathbb{R}^{d_c}$. This vector is additionally passed through a batch normalization. For injection into the U-Net backbone of the latent diffusion model, it is reshaped to match the spatial dimensions of the features maps $(H_{fm},W_{fm})$ at each resolution $e_c \rightarrow (d_c,H_{fm},W_{fm})$, which is achieved by repeating $e_c$ for every pixel in the feature map. The final reshaped tensor $e_c \in \mathbb{R}^{(b,d_c,H_{fm},W_{fm})}$ also includes the batch size $b$ and is concatenated with the feature maps before being passed into the first residual block (ResBlock) of each resolution level in the U-Net. 

Our LDM follows the architecture used in \citep{rombachHighResolutionImageSynthesis2022} and \citep{FisherPlantLDM2022}. However, we remove cross-attention for efficiency. The input data $x$ (images in our case) is passed through the pretrained VAE-encoder $\mathcal{E}$ from Stable Diffusion 2.1 \citep{rombachHighResolutionImageSynthesis2022,vonplatenDiffusersStateoftheartDiffusion2024} to obtain the latent representation $z=\mathcal{E}(x)$. The U-Net backbone $\hat{\epsilon}_{\theta}={\epsilon}_{\theta}(z_t,t,e_c)$ operates on the noisy latent $z_t$ and is conditioned on both the timestep $t$ and the processed conditioning vector $e_c$. It is trained to predict the noise $\epsilon$ added at timestep $t$. The loss for training the LDM is calculated as the MSE between the true added noise $\epsilon$ and the predicted noise $\hat{\epsilon}$ \citep{ho2020denoisingdiffusionprobabilisticmodels}:
\begin{equation}
\footnotesize
    \mathcal{L}_{\text{LDM}}=\mathbb{E}_{t,z,\epsilon}[\parallel \epsilon-\epsilon_{\theta}(z_t,t,e_c)\parallel^2_2].
\end{equation}
Following the denoising, we employ the pretrained Stable Diffusion VAE-decoder $\mathcal{D}(z_c)$ to transform the conditioned latent $z_c$ back into the image space.

\section{Experiments}
Our experiments investigate the performance of our proposed masked conditioning approach on sparse and mixed-type conditioning information. We provide implementations for point-cloud- and image-data.

\subsection{Masked Conditioning for VAE}
For point-cloud data, we apply the masked conditioning to a VAE-architecture (\Cref{subsec:mcVAE}). Our experiments are conducted using two datasets. We use GeoBiked \citep{muellerGeoBiked2024}, where bicycle geometries are represented as 12 characteristic reference points and are annotated with seven categorical and one numerical feature. The dataset contains a total of 4355 samples. Our second dataset is proprietary and contains 782 silhouettes of passenger vehicles, represented as structured point-clouds with 21 reference points describing the geometry, as well as six categorical and one numerical features.\\

\noindent\textbf{Model Configurations and Performance.}
For the model trained on the GeoBiked, we optimize the baseline architecture by conducting hyperparameter tuning in two stages. We use the Optuna framework \citep{optuna_2019} for an exploration of the hyperparameter space, followed by a focused grid search to refine critical parameters. We conduct 1000 optimization steps with Optunas TPESampler and test another 900 hyperparameter configurations in the subsequent grid search. The parameters of the final model configuration are summarized in \Cref{tab:VAE_Tab}.

\begin{table}[h]
\caption{Model parameters of the VAE trained with masked conditioning with the BIKED and the passenger vehicle dataset after the two-stage hyperparameter tuning.}
\scriptsize
\renewcommand{\arraystretch}{1.3}
\centering
\begin{tabularx}{\columnwidth}{UUU}
\toprule
Parameter & GeoBIKED Dataset \citep{regenwetterBIKEDDatasetComputational2021} & Vehicle Dataset \\
\midrule
 Reference Keypoints (Quantity) & 12 & 21\\
 Categorical Conditions (Quantity) & 7 & 6\\
 Numerical Conditions (Quantity) & 1 & 1\\
 \midrule
 Reference Keypoints Embedding Dimension & 203 & 151\\
 Conditioning Embedding Dimension & 11 & 19 \\
 Batch Size & 140 & 55\\
 Epochs & 393 & 324\\
\bottomrule
\end{tabularx}
\label{tab:VAE_Tab}
\end{table}

\begin{figure}[h]
\begin{minipage}{0.48\textwidth}
    \centering
    \includegraphics[height=5cm,keepaspectratio]{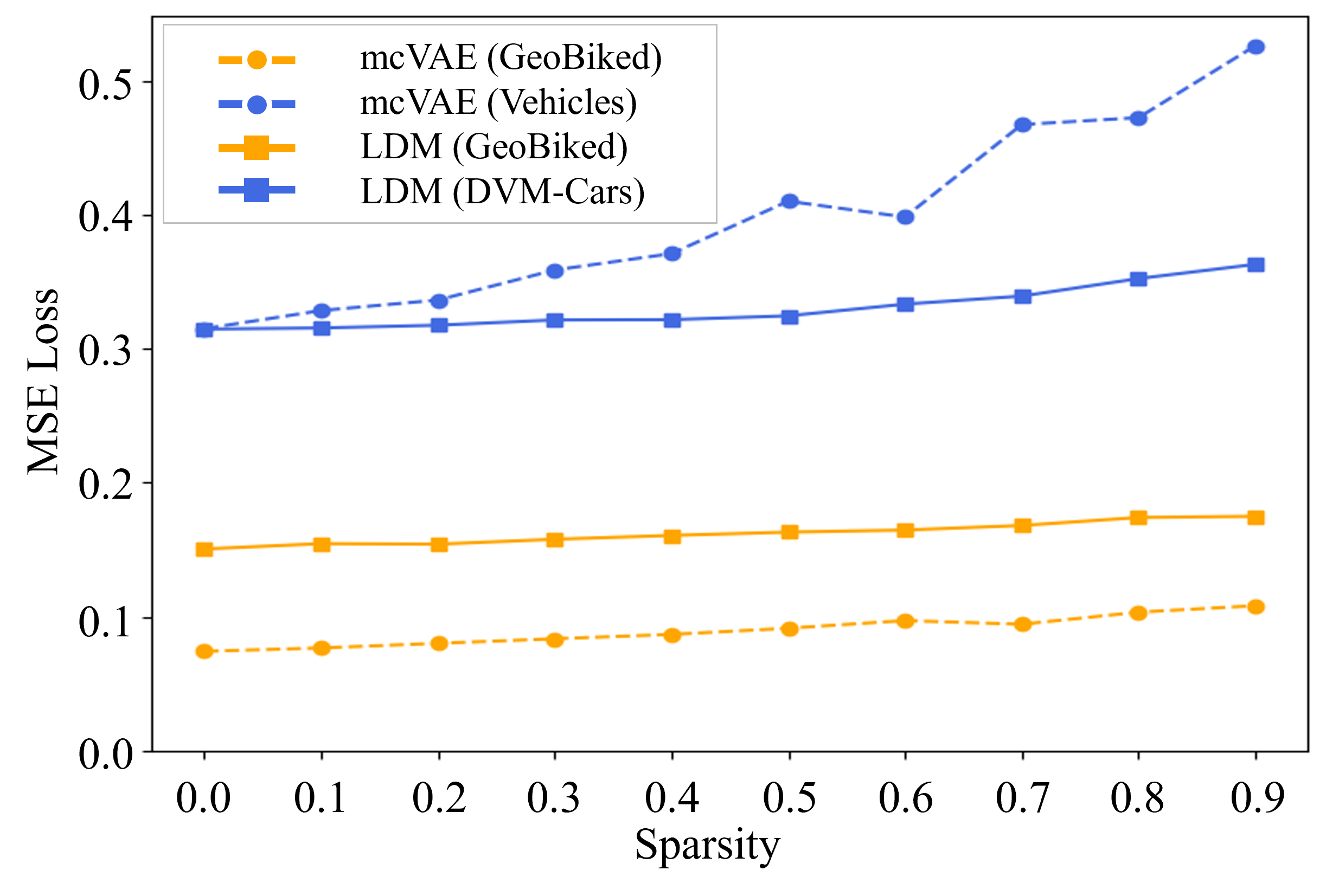}
\end{minipage}%
\hfill
\begin{minipage}{0.48\textwidth}
    \centering
    \includegraphics[height=5cm,keepaspectratio]{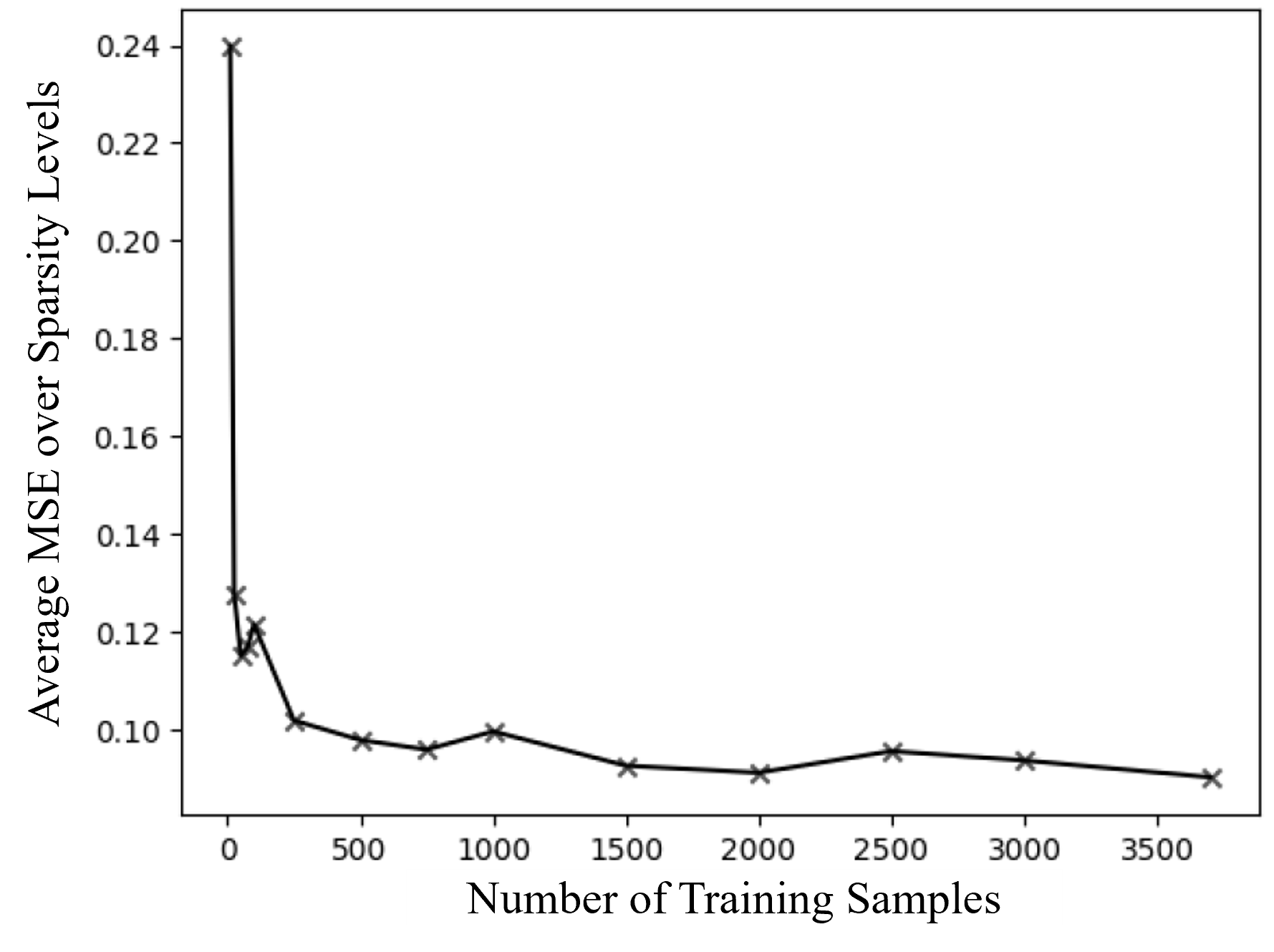}
\end{minipage}%
\caption{\textbf{Left:} MSE for the VAEs and LDMs trained on the GeoBIKED, vehicles and quality checked DVM-Car subset datasets for increasing levels of sparsity in the conditions. The sparsity levels are kept constant for each training run. \textbf{Right:} Mean MSE over sparsity levels over the number of samples in the training dataset (BIKED dataset).}
\label{fig:MSEs}
\end{figure}

For both datasets, we train a model with the hyperparameters stated in \Cref{tab:VAE_Tab} on conditioning sparsity levels, ranging from 0.0 to 0.9. A sparsity level of 0.1 indicates that 10\% of the conditioning data is randomly masked during training, simulating incomplete or missing information. 
The Mean-Squared-Error (MSE) measures the accuracy of the generated reference points against the ground-truth in the test dataset. As expected, it increases as the sparsity level rises (\Cref{fig:MSEs}). For GeoBiked, we achieve an average MSE of 0.0895 across all sparsity levels. The approach robustly reconstructs the reference points even with high levels of missing conditioning information. The model trained on the smaller passenger vehicle dataset exhibits a higher mean MSE of 0.3985. This can be explained by the dataset's reduced sample size and the inherent ambiguity in the vehicle geometries, which makes the reconstruction task more challenging.\\

\noindent\textbf{Dataset Size.}
To assess the impact of dataset size on the performance of the masked conditioning approach, we conduct experiments on the GeoBIKED dataset, for which the larger size allows for a detailed analysis across various dataset sizes. We train models on subsamples of the dataset, with sizes ranging from 10 to 3,700 samples, and evaluate the models across multiple sparsity levels (from 0.0 to 0.9). The mean MSE for different dataset sizes, averaged over all sparsity levels, is summarized in \Cref{tab:VAE_Model_Size}.

For very small datasets, e.g. 10 or 50 samples, higher sparsity results in better accuracy. At small sample sizes the model has insufficient data to generalize effectively when too much conditioning information is introduced. In these cases, removing some conditioning data allows the model to focus on the more general distribution of the reference points rather than overfitting to the limited number of conditions. As the dataset size increases, however, more conditioning information leads to better accuracy, as the model is able to effectively utilize the available data to condition the generative process.

We observe that a dataset size of approximately 500 samples appears to be the threshold beyond which further increases in dataset size provide minimal improvements in accuracy (\Cref{fig:MSEs}). While there are improvements in MSE with larger datasets (up to 3,500 samples), the improvements are marginal. These findings suggest that our masked conditioning approach is data-efficient, requiring relatively small datasets to achieve competitive accuracy, particularly when the conditions are sparse.\\

\begin{table}[h]
\caption{Model Performance (MSE) on different sizes of the GeoBIKED dataset over training sparsities using the mcVAE. Each column represents a sparsity level. Training has been conducted with constant sparsity for each level.}
\scriptsize
\renewcommand{\arraystretch}{1.3}
\centering
\begin{tabularx}{\columnwidth}{WYYYYYY}
\toprule
Dataset Size & 0.0 & 0.2 & 0.4 & 0.6 & 0.8 & \textbf{Mean} \\
\midrule
10 & 0.3212 & 0.2810 & 0.2395 & 0.2106 & 0.1798 &  \textbf{0.2398}\\
100 & 0.1201 &  0.122 & 0.1187  & 0.122 & 0.1239  &  \textbf{0.1217}\\ 
1000 & 0.090 &  0.0929 & 0.0990  & 0.1045 & 0.1088 &  \textbf{0.0998}\\
2000  & 0.0767 &  0.0820 & 0.0892 & 0.0985 &  0.1035 &  \textbf{0.0914}\\
3000 & 0.0759 & 0.0835 & 0.0910 &  0.1029 & 0.1079 &  \textbf{0.0939}\\
\bottomrule
\end{tabularx}
\label{tab:VAE_Model_Size}
\end{table}

\noindent\textbf{Sparsity Schedule.}
In addition to evaluating models trained on fixed sparsity levels, we conduct experiments on various sparsity scheduling strategies during training. These experiments aim to determine if dynamic sparsity schedules could enable models to generalize across a wider range of sparsity levels during inference. Four different sparsity schedules are tested: constant, linear, step-wise, and exponential. For the non-constant schedules we test increasing and decreasing sparsities.

The results for both datasets are summarized in \Cref{tab:VAE_Sparsity_0.5_Experiments}. Overall, no single sparsity schedule consistently outperforms all others across every inference sparsity level. Low sparsity levels during training generally lead to good performance across all sparsity levels during inference. High levels of sparsity in the conditioning information is only beneficial for high levels of sparsity at inference time. 

For the GeoBIKED dataset, using a sparsity schedule provides no significant benefit over a constant sparsity level. If a sparsity schedule is used, increasing the sparsity throughout training is superior to decreasing it. If a conditioning sparsity of 0.5 is used in training with a constant sparsitly schedule to simulate sparse training data (\Cref{tab:VAE_Sparsity_0.5_Experiments}), this seems to improve the generation even if the model is conditioned on all inputs during inference. When the generation is conditioned on an arbitrary sparsity of inputs (i.e. inference sparsity levels are 0.0 to 0.9), an exponential schedule performs marginally better.
For the vehicle dataset, a decreasing sparsity is superior. When only the training runs with a sparsity of at least 0.5 are evaluated (\Cref{tab:VAE_Sparsity_0.5_Experiments}), keeping the sparsity constant yields the best performance. 

\begin{table}[h]
\caption{Comparison of the MSE of the mcVAE on the two datasets when the minimum sparsity level in training is 0.5. For GeoBiked, all shown configurations are trained with increased sparsity $(0.5 \sim 0.6)$, while for the vehicles sparsity decreased $(0.6\sim0.5)$.}
\scriptsize
\centering
\begin{tabularx}{\linewidth}{ZcZZZZ}
\toprule
Dataset & Inference Sparsity Level & Constant & Linear & Stepwise & Exponential\\
\midrule
\multirow{2}{*}{\textit{GeoBIKED}}
    & \textit{0.0} & \textbf{0.0766 (0.5)} & 0.0771  & 0.0775  &  0.0773  \\   
    & \textit{0.0-0.9} & 0.0898 (0.5) & 0.0896 & 0.0900 & \textbf{0.0895}\\
\midrule
\multirow{2}{*}{\textit{Vehicles}}
    & \textit{0.0} & \textbf{0.3941 (0.5)} & 0.4317 & 0.4310 & 0.4377 \\
    & \textit{0.0-0.9} & \textbf{0.4791 (0.5)} & 0.5196 & 0.5202 & 0.5237 \\
\bottomrule
\end{tabularx}
\label{tab:VAE_Sparsity_0.5_Experiments}
\end{table}

\subsection{Masked Conditioning for LDM}
Our masked conditioning approach is also integrated into latent diffusion models, which are a state-of-the-art method for generating images. Large-scale diffusion models for high-quality natural image synthesis are trained on hundreds of millions of images. They are infeasible to train from scratch for domain specific applications. However, latent diffusion models for special, limited applications can be trained from scratch, only requiring moderate amounts of compute. We train a latent diffusion architecture, as described in \Cref{subsec:mcLDM}, using our masked conditioning approach. 

In our experiments, we use two image datasets. We again used GeoBIKED \citep{muellerGeoBiked2024}, which provides image representations of the bicycles in addition to categorical and numerical features. We also use the DVM-Car dataset \citep{Huangdvmcar2022}. The quality-checked subset of DVM contains 67k front view images of passenger vehicles. For annotations, we use 14 categorical and 10 numerical features.
Training our LDM is inherently more expensive than training the VAE-architecture. Therefore, we do not conduct sparsity scheduling experiments but focus on evaluating the model performance in terms of quality of the generated images. We set the sparsity to linearly increase from 0.1 to 0.25 during training for both models. This should result in good generalization capabilities for all sparsity levels in inference.\\

\begin{figure}
\centering
\includegraphics[width=\linewidth]{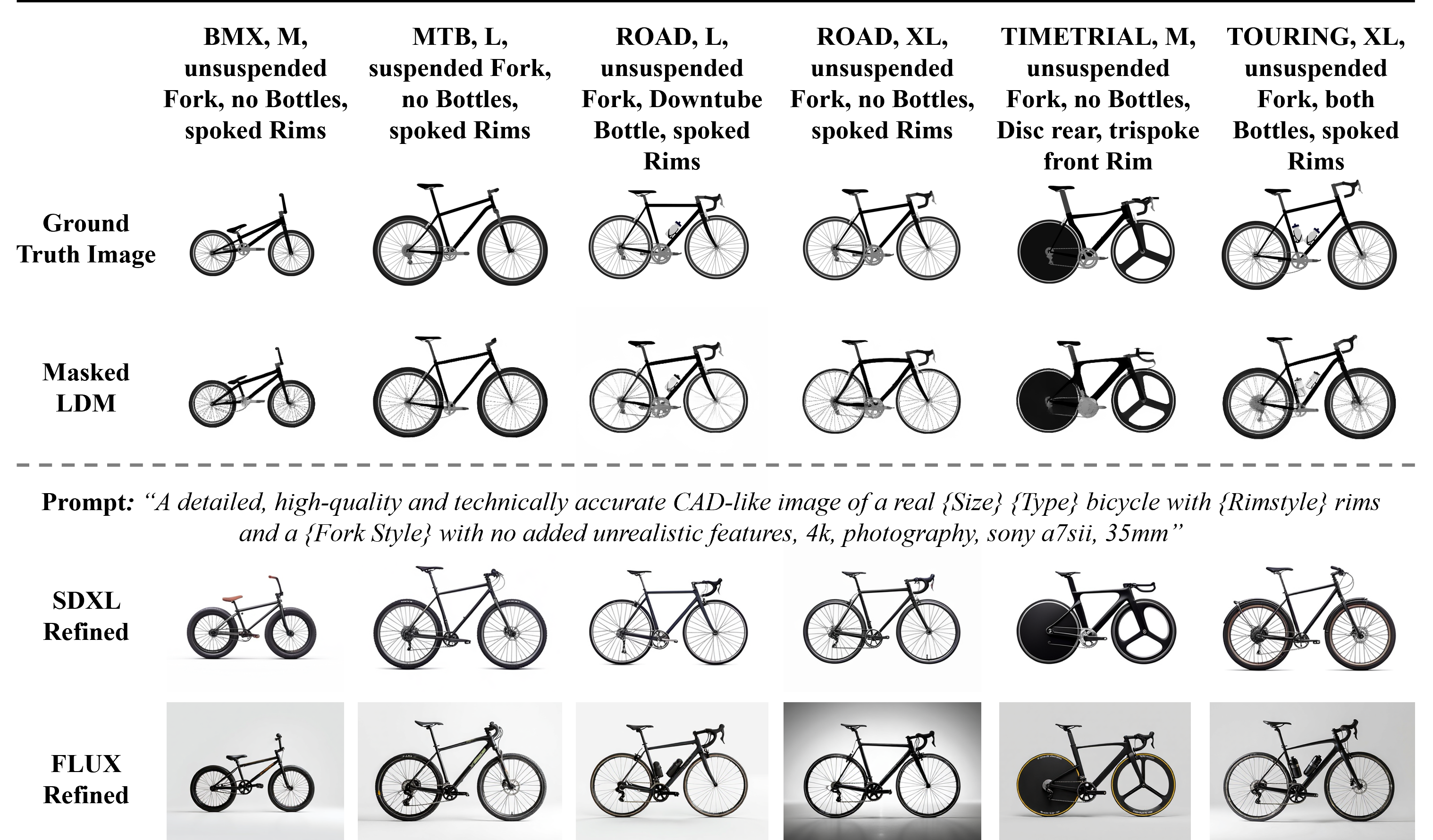}
\caption{Qualitative results of reconstructing images from the GeoBiked dataset. The mcLDM is conditioned with the same inputs as the ground truth image is labeled. For the refinement, the mcLDM-generated image is passed to the model as input, together with the prompt. Best viewed when zoomed in.}
\label{fig:Refinement}
\end{figure}

\noindent\textbf{Model Configurations and Performance.}
Our mcLDM-architecture is based on the PlantLDM  \citep{FisherPlantLDM2022}, which is a simplified unconditional implementation of Stable Diffusion \citep{rombachHighResolutionImageSynthesis2022}. In the U-Net, we increase the starting channel dimension to 64 for the GeoBIKED data and to 128 for the DVM-Car data. The U-Net levels are set to 4 and the number of attention heads is increased to 8 for both models. We use the pretrained Stable Diffusion 2.1 VAE for encoding the images. For the model trained on GeoBiked, we employ a batch size of 32. The DVM-Car model is trained with a larger batch size of 128. The learning rate is $1e{-4}$ for both. Some results of the conditional generation on both datasets are visualized in the upper rows of \Cref{fig:Refinement}.

The experiments with our mcLDM demonstrate the utility of our masked conditioning mechanism. We emphasize that large-scale, state-of-the-art LDMs achieve notably higher image fidelity. Given the relatively small size of the dataset, the lack of specific optimization and the moderate computational cost (approx. 30 hours on a single A6000), we deem the quality of the generated images acceptable to investigate the capability of our method to handle sparse and mixed conditioning information. Further, we will later discuss how to obtain high fidelity generations without additional training.

Considering the influence of different sparsity levels at inference time, we observe that the accuracy of the generation deteriorates slightly with increasing sparsity (\Cref{fig:MSEs}). This is coherent with the results of our experiments with the VAE-architecture. While the MSE increases for a higher inference sparsity, we are still able to generate feasible images from few input conditions. Both LDMs are trained with a masking schedule that simulates sparse conditioning data. When tested on various sparsity levels at inference time, we observe that the reconstruction of ground-truth samples works reasonably well. In \Cref{fig:Refinement}, the upper rows show results of generated samples with the mcLDM compared to the ground-truth images when using the same conditioning inputs.\\

\noindent\textbf{Image Refinement with Pretrained Models.}
We want to highlight the possibility of training small-scale conditional models that act as generative priors on domain-specific data which align well with the conditioning data while large-scale models for image generation achieve significantly higher image fidelity and realism. In an attempt to improve efficiency while retaining image quality in practical applications, we propose to combine both approaches. With our small-scale LDM it is possible to generate a low fidelity representation of the technical object that adheres to the proposed conditions. This image can then be refined or visualized in realistic scenes through the utilization of pretrained architectures.

Stable Diffusion XL (SDXL) \citep{podellSDXLImprovingLatent2023} provides a model for image-to-image refinement. In \Cref{fig:Refinement}, we show some qualitative results for both mcLDMs. For this refinement, we employ the SDXL-refiner with a guidance-scale of 7.5 and a diffusion-strength of 0.5. We additionally employ FLUX \citep{flux2023}, a state-of-the-art Diffusion Transformer \citep{peebles2023Scalable, esser2024scalingrectifiedflowtransformers}, for image refinement. We again use the image generated by our mcLDM as conditioning input, together with the prompt that describes the specific features. The image refinement is conducted for 50 inference steps with a guidance scale of 30.0.

While using only mcLDM results in better alignment with ground truth images (lower MSE and LPIPS, higher CLIP-similarity and SSIM scores), the refinement by SDXL and FLUX introduces a significant enhancement in photorealism and perceived image quality. The hybrid approach showcases the potential of using small-scale, domain-specific generative models as priors, refined by general-purpose large-scale architectures. Hence, we can leverage the controllability and efficiency of smaller models while benefitting from the expressive power of large pretrained architectures. 
\begin{figure}[h]
\centering
\includegraphics[width=\linewidth]{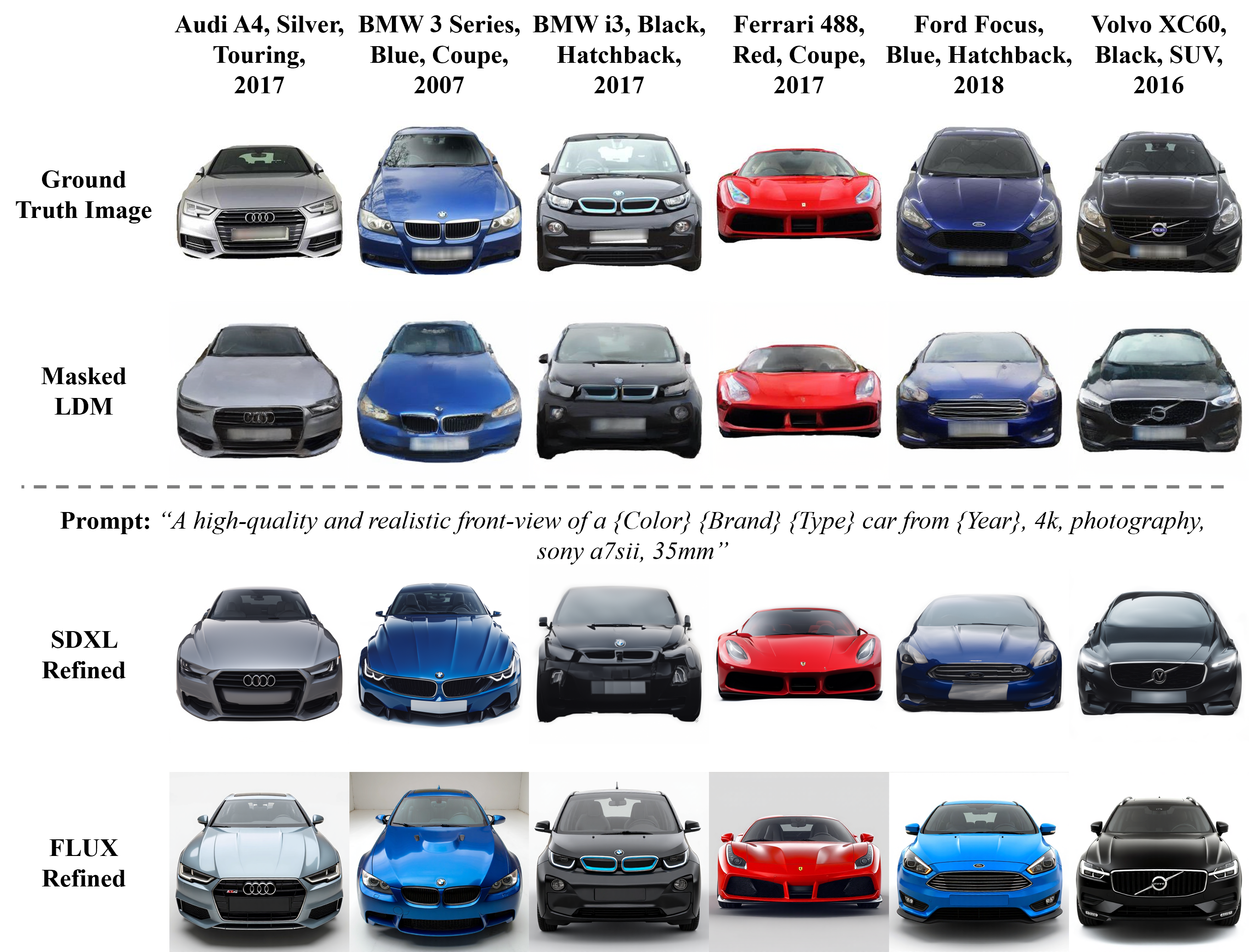}
\caption{Qualitative results of reconstructing images from the DVM-Car dataset. The mcLDM is conditioned with the same inputs as the ground truth image is labeled. For the refinement, the mcLDM-generated image is passed to the model as input, together with the prompt. Best viewed when zoomed in.}
\label{fig:Refinement_DVM}
\end{figure}

\begin{table}[h]
\centering
\caption{Quantitative results on the test samples from \Cref{fig:Refinement} against the ground truth image. The results are averaged over the visualized test samples.}
\scriptsize
\renewcommand{\arraystretch}{1.3}
\centering
\begin{tabularx}{\columnwidth}{WWWWW}
\toprule
Image samples & MSE $\downarrow$ & CLIP-similarity \citep{hessel2022clipscorereferencefreeevaluationmetric} $\uparrow$ & SSIM \citep{WangSSIM} $\uparrow$ & LPIPS \citep{LPIPSzhang2018unreasonableeffectivenessdeepfeatures} $\downarrow$ \\
\midrule
mcLDM (GeoBiked) & 0.0269 & 0.9854 & 0.8583 & 0.0681  \\
SDXL-Refined & 0.0266 & 0.9634 & 0.8502 & 0.0762  \\
FLUX-Refined & 0.0870 &  0.9075 & 0.7868  & 0.1987\\
\midrule
mcLDM (DVM) & 0.0678 & 0.9010 & 0.5862 & 0.3642 \\
SDXL-Refined & 0.0730 & 0.8709 & 0.5703 & 0.4067 \\
FLUX-Refined & 0.1000 & 0.8575 & 0.4796 & 0.4695 \\
\bottomrule
\end{tabularx}
\label{tab:VAE_Model_Size}
\end{table}


\FloatBarrier
\section{Conclusion}
Our novel masked conditioning method is tailored for training generative models on engineering datasets that are often sparse, and composed of mixed numerical and categorical conditions demonstrates versatility across multiple tasks, including both point clouds and image data.  Sparsity scheduling allows the generative model to be trained on sparse data while maintaining its generative capabilities. However, the approach has some limitations. The models currently lack extrapolation capabilities to completely novel conditions and likely require additional data for better generalization. The performance of the mcLDM is constrained by the pretrained VAE used, with finetuning being both unstable and resource-intensive. Further, handling more complex conditions, such as high-dimensional categorical inputs or image and text-based conditioning, remains a challenge. 

We show that small-scale generative models can act as domain-specific prior generators to large pretrained models, enabling customized generation and high image quality. This modular approach not only balances efficiency with quality but also allows for seamless integration of more advanced generative architectures as they become available.

\printcredits

\section*{Declaration of competing interest}
The authors declare that they have no known competing financial interests or personal relationships that could have appeared to
influence the work reported in this paper.

\section*{Data Availability}
Dataset and Code are publicly available and can be found under:\\
\url{https://anonymous.4open.science/r/Masked_Conditioning-E2BB}.

\section*{Declaration of generative AI and AI-assisted technologies in the writing process}

During the preparation of this work the author(s) used ChatGPT in order to improve readability of the manuscript. After using this tool/service, the author(s) reviewed and edited the content as needed and take(s) full responsibility for the content of the publication.

\bibliographystyle{cas-model2-names}

\bibliography{Masked_VAE}

\end{document}